%% file: main.tex
\documentclass[letterpaper]{article} 

\usepackage[preprint]{neurips_2023}
\usepackage{times}  
\usepackage{helvet}  
\usepackage{courier}  
\usepackage[hyphens]{url}  
\usepackage{graphicx} 
\usepackage{natbib}  
\usepackage{caption} 
\usepackage{algorithm}
\usepackage{algorithmic}
\usepackage{fix-cm}
\usepackage{layout}
\usepackage{csquotes}
\usepackage{subcaption}
\usepackage{newfloat}
\usepackage{listings}
\usepackage{appendix}
\DeclareCaptionStyle{ruled}{labelfont=normalfont,labelsep=colon,strut=off} 
\floatstyle{ruled}
\newfloat{listing}{tb}{lst}{}
\floatname{listing}{Listing}
\usepackage{xcolor}

\usepackage{amsfonts}
\usepackage{amsmath}
\usepackage{amsthm }
\usepackage{dsfont}
\usepackage[noend,algo2e,noline,ruled,linesnumbered]{algorithm2e}
\usepackage{multirow}
\usepackage{booktabs}
\usepackage{nicefrac}

\newcommand{\indicator}{\mathds{1}}
\title{Model Agnostic Explainable Selective Regression via Uncertainty Estimation}
\begin{document}
\newcommand{\dan}[1]{\textcolor{violet}{[D: #1]}}
\newcommand{\carlos}[1]{\textcolor{blue}{[C: #1]}}
\newcommand{\andrea}[1]{\textcolor{orange}{[A: #1]}}
\newcommand{\plugin}{\textsc{PlugIn}}
\newcommand{\doubtvar}{\textsc{DoubtVar}}
\newcommand{\doubtint}{\textsc{DoubtInt}}
\newcommand{\scross}{\textsc{SCross}}
\newcommand{\mapie}{\textsc{Mapie}}
\newcommand{\gold}{\textsc{GoldCase}}

\author{%
Andrea Pugnana\\  
Scoula Normale Superiore\\
Italy  \\
\And
Carlos Mougan \\
University Southampton\\
United Kingdom
\And 
Dan Saattrup Nielsen\\
Alexandra Institute\\ Denmark\\
}
\maketitle

\input{content/intro}
\input{content/relatedwork}
\input{content/methodology}
\input{content/experiments}
\input{content/conclusions}
\bibliography{biblio}
\bibliographystyle{apalike}
\appendix
\setcounter{table}{0}
\renewcommand{\thetable}{A\arabic{table}}

\input{appendixV2}

\end{document}

%% file: content/intro.tex
\begin{abstract}
    With the wide adoption of machine learning techniques, requirements have evolved beyond sheer high performance, often requiring models to be trustworthy.
    A common approach to increase the trustworthiness of such systems is to allow them to refrain from predicting.
    Such a framework is known as selective prediction. While selective prediction for classification tasks has been widely analyzed, the problem of selective regression is understudied. This paper presents a novel approach to selective regression that utilizes model-agnostic non-parametric uncertainty estimation. Our proposed framework showcases superior performance compared to state-of-the-art selective regressors, as demonstrated through comprehensive benchmarking on 69 datasets. Finally, we use explainable AI techniques to gain an understanding of the drivers behind selective regression. We implement our selective regression method in the open-source Python package \texttt{doubt} and release the code used to reproduce our experiments.
\end{abstract}

\section{Introduction}
Selective prediction allows machine learning systems to refrain from forecasting, adding the option to abstain whenever the risk of mispredicting is too high. 
This is an appealing feature for contexts where making wrong predictions can produce relevant harm, such as healthcare or finance.
Under this framework, we can distinguish between selective classification and selective regression, similarly to the supervised learning paradigm.

While selective classification has received considerable attention in the past decade, the problem of selective regression remains relatively understudied. Existing approaches for selective regression primarily rely on either deep-learning-based methods~\cite{DBLP:conf/icml/GeifmanE19,DBLP:conf/ijcnn/JiangZW20} or on the ability to estimate the conditional variance function \cite{DBLP:conf/nips/ZaouiDH20}. However, due to the limited scope of empirical evaluations in current state-of-the-art methods, there is a lack of practical insights into the effectiveness and applicability of existing techniques, hindering the development of robust and reliable selective prediction models.

This paper aims to bridge this gap by introducing a
novel model-agnostic method for selective regression using non-parametric bootstrap estimators. Our proposed approach offers a robust and reliable solution for abstaining from predicting
in regression tasks.

To validate the effectiveness of our selective regression methodology, we conduct extensive empirical evaluations on a set of 69 tabular datasets. This larger-scale benchmarking allows us to gain insights into the performance and generalizability of our proposed approach. Additionally, we incorporate explainable AI methodologies to understand better the factors contributing to the rejection of predictions. By identifying the sources of rejection, we enhance the interpretability of selective regressors, enabling the system users to have accountability over such systems. Our main contributions are:
\begin{itemize}

\item a state-of-the-art selective regression method using non-parametric bootstrap to estimate uncertainty;
    
\item an extensive empirical evaluation of selective regression techniques on 69 tabular datasets;

\item the application of explainable AI methods to identify the sources of rejection/prediction.
\end{itemize}


%% file: content/relatedwork.tex
\section{Related Work}\label{sec:related_work}

\paragraph{Prediction with a Reject Option} The idea to allow a machine learning model to abstain in the prediction stage dates back to the 1970s \cite{DBLP:journals/tit/Chow70}, when it was introduced for the classification task.

We can distinguish two main frameworks that allow us to learn such a pair:
Learning to Reject (LtR)~\cite{DBLP:journals/tit/Chow70,Hendrickx2021} and Selective Classification (SC) \cite{DBLP:journals/jmlr/El-YanivW10}.
The former (LtR) requires one to define a class-wise cost function that penalizes mispredictions and rejections~\cite{Herbei06,DBLP:conf/nips/CortesDM16,DBLP:journals/prl/Tortorella05,DBLP:conf/isbi/CondessaBCOK13}.
The latter (SC) requires instead one to pre-define either a target coverage $c$ or a desired target risk $e$ to achieve.
Depending on this choice, a selective classifier can be learnt by either minimizing the risk given the target coverage \cite{DBLP:conf/icml/GeifmanE19,DBLP:conf/nips/LiuWLSMU19,DBLP:conf/nips/Huang0020,PugnanaRuggieri2023a,PugnanaRuggieri2023b,feng2023towards} or by maximizing coverage given a target risk \cite{DBLP:conf/nips/GeifmanE17,DBLP:conf/aistats/GangradeKS21}.
The conditions for equivalence between SC and LtR are studied in ~\cite{DBLP:conf/icml/FrancP19}.

A few related works have focused on embedding regressors with the reject option.~\citet{DBLP:conf/icml/GeifmanE19} proposes a deep learning method called $\textsc{SelNet}$ that, given a target coverage, jointly trains the regressor and the selection function.
\citet{DBLP:conf/ijcnn/JiangZW20} propose the usage of deep ensembles for selective regression in weather forecasting.
On the other hand, \citet{DBLP:conf/nips/ZaouiDH20} rely on the \textit{Plug-In principle} to estimate the optimal selective regressor in a model-agnostic way. 

\paragraph{Uncertainty Estimation}
Traditionally, uncertainty has been closely associated with standard probability and probabilistic predictions in the statistical realm. However, in the field of machine learning, new challenges have arisen, such as trust, robustness, and safety, necessitating the development of novel methodological approaches~\cite{DBLP:journals/ml/HullermeierW21}.

A widely adopted method for uncertainty estimation in machine learning is model averaging~\cite{kumar2012bootstrap, DBLP:conf/icml/GalG16, DBLP:conf/nips/Lakshminarayanan17}. This approach has proven successful in various applications, including monitoring model degradation~\cite{mougan2023monitoring}, detecting misclassifications~\cite{DBLP:conf/nips/RenLFSPDDL19}, and addressing adversarial attacks~\cite{DBLP:conf/cvpr/HendrycksZBSS21}, among others.

In our research, we focus on applying uncertainty to selective regression. Specifically, we extend existing model-agnostic uncertainty frameworks to calculate a confidence function for cases where we abstain from making predictions~\cite{kumar2012bootstrap, mougan2023monitoring}. Additionally, we explore and compare another approach known as conformal prediction, as presented in the work by~\citet{DBLP:conf/nips/KimXB20}.

\paragraph{Explainable Reject Option}
Explainable AI methods \cite{DBLP:journals/csur/GuidottiMRTGP19} are increasingly used to overcome the difficulty of interpreting AI outputs. Adding explanations to rejections allows for characterizing the areas where the predictor is not confident enough~\cite{Pugnana2023}. To the best of our knowledge, the current literature on explaining selective function is limited. \citet{DBLP:conf/ijcci/ArteltBVH22} propose local model agnostic methods to perform such a task;
\citet{DBLP:conf/esann/ArteltVH22} use counterfactual explanations to explain the reject option for Learning Vector Quantization (LVQ) algorithms.

In this paper, similarly to ~\cite{mougan2023monitoring}, we try to explain the reasons behind the reject option by learning a binary classifier that predicts acceptance/rejection of the original estimator and evaluating its decisions using Shapley Values~\cite{shapTree}.


%% file: content/methodology.tex
\section{Methodology}

\subsection{Selective Regression Framework}

Let $X$ and $Y$ be random variables taking values in $\mathcal{X}\subseteq\mathbb{R}^d$ and $\mathcal{Y}\subseteq\mathbb{R}$, respectively. A \textbf{predictor} is a function $f\colon\mathcal{X}\to\mathcal{Y}$ and a \textbf{selection function} is a predictor taking values in $\{0,1\}$. We further define, for a predictor $f$ and a selection function $s$, an associated \textbf{selective predictor} $f_s$ as 
\begin{equation}
    f_s(X) = \begin{cases}
        f(X) & \text{if } s(X)=1\\
        \emptyset & \text{otherwise},
    \end{cases}
\end{equation}

where $\emptyset$ denotes abstaining from predicting. However, the direct estimation of $s$ can be challenging: for instance, $s$ is not differentiable. Therefore, the selection function can be relaxed by considering an associated \textbf{confidence function}\footnote{A good confidence function $c_f$ should rank instances based on descending (user-defined) loss $l$, i.e. if $c_f(\mathbf{x}_i) \leq c(\mathbf{x}_j)$ then $l(f(\mathbf{x}_i),y_i) \geq l(f(\mathbf{x}_j),y_j)$.} $c_f\colon\mathcal{X}\to\mathbb R$, sometimes called soft selection~\cite{DBLP:conf/nips/GeifmanE17}, that measures how likely the predictor $f$ is correct. We can then set a threshold $\tau\in\mathbb R$ that defines the minimum confidence for providing a prediction, yielding the associated selection function $x\mapsto\indicator(c_f(x)>\tau)$.

\begin{figure}[ht]
    \centering
    \includegraphics[width=1.0\linewidth]{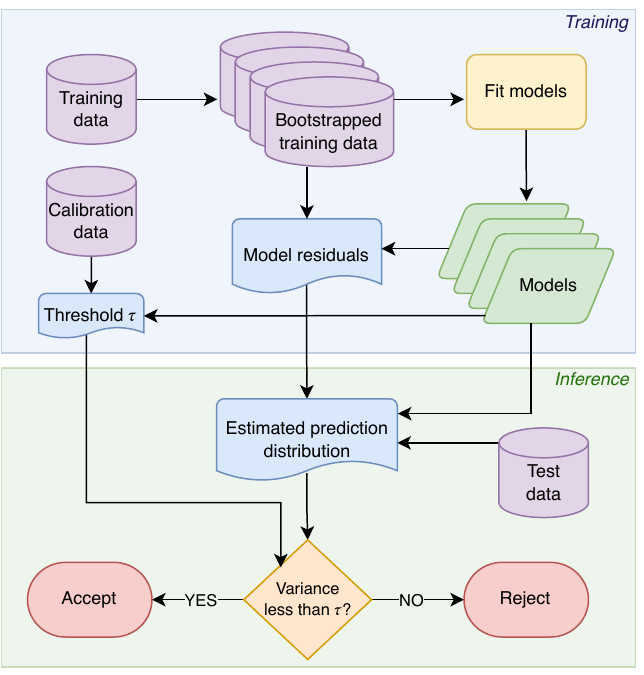}
    \caption{Our \doubtvar{} selective regression method. Our \doubtint{} method is identical, aside from replacing variance with the difference between the 97.5\% quantile and the 2.5\% quantile.}
    \label{fig:diagram-selective-regression}
\end{figure}

\begin{figure}[ht]
    \centering
    \includegraphics[width=1.0\linewidth]{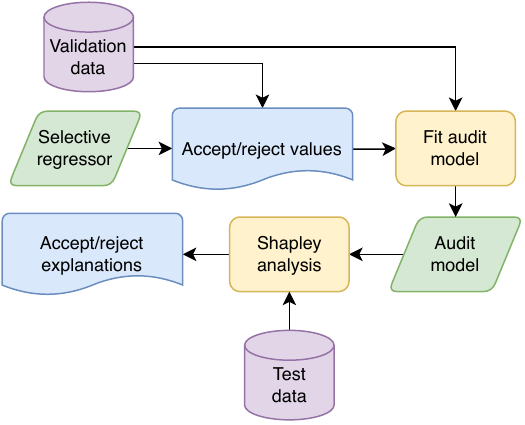}
    \caption{Our explainable selective regression method.}
    \label{fig:diagram-xai}
\end{figure}

Two important measures are associated with the selective predictor $f_s$. The \textbf{coverage} $\texttt{cov}(s) := \mathbb{E}[s(X)]$, being the expected mass probability of the non-rejected region, and the \textbf{selective risk}
\begin{equation}
    \texttt{risk}(f,s) := \mathbb{E}[l(f(X),Y)\mid s(X)=1],
\end{equation}

being the expected error over accepted instances, where $l\colon\mathcal{Y}\times\mathcal{Y}\to\mathbb{R}$ is a user-defined loss function. Given a target coverage $\alpha$, an \textbf{optimal selective predictor} $f_s^*$, parameterised by $\theta_f$ and $\theta_s$, is a solution to
\begin{equation}
    \label{eq:spp}
    \min_{\theta_f,\theta_g}\texttt{risk}(f_{\theta_f},s_{\theta_s})\quad\text{such that}\quad\texttt{cov}(s_{\theta_s})\geq\alpha.
\end{equation}

\citet{DBLP:conf/nips/ZaouiDH20} showed that the optimal selective predictor under the mean squared error loss function is obtained by
\begin{equation}
    \label{eq:optimal-selective-predictor}
    f_g^*(x) = \begin{cases}
        f^*(x) & \text{if }\sigma^2(x)<\tau_\alpha\\
        \emptyset & \text{otherwise},
    \end{cases}
\end{equation}

where $f^*(x):= \mathbb{E}[Y\mid X=x]$ is the regression function, $\sigma^2(x) := \mathbb{E}[(Y-f^*(x))^2\mid X=x]$ is the conditional variance function and $\tau_\alpha$ is such that \begin{equation}
    \texttt{cov}(\indicator(\sigma^2(X) < \tau_\alpha)) = \alpha.
\end{equation}

In other words, the optimal selective predictor associated with a target coverage $\alpha$ forecasts whenever the variance is below $\tau_\alpha$.

Finally, we call \textit{calibration} the post-training procedure of estimating the threshold $\tau_\alpha$ in (\ref{eq:optimal-selective-predictor}) for the target coverage $\alpha$ specified in (\ref{eq:spp}).
This is generally done by estimating the ~$1-\alpha$ quantile of the confidence
over a held-out calibration dataset \cite{DBLP:conf/icml/GeifmanE19}.

\subsection{Uncertainty Estimation as a Confidence Function}
\label{sec:our-method}

In this paper, we propose using uncertainty estimation techniques as confidence functions in the sense described in the previous section. We will use a variation of the state-of-the-art \texttt{doubt} uncertainty estimation method \cite{mougan2023monitoring}, which we will briefly describe here. We assume the following relationship between the true values and the model predictions:

\begin{equation}
    Y = f(X) + \varepsilon(X)
\end{equation}

Here $\varepsilon(X)$ is the combined noise function, which can be split into \textit{model variance noise}, \textit{observation noise} and \textit{model bias}. 
As the goal is to estimate $\varepsilon(X)$, 
\citet{mougan2023monitoring} build a set $C(X)$ consisting of bootstrapped values whose distribution estimates the distribution of $\varepsilon(X)$. Indeed, as this method does not change the asymptotic properties of the prediction interval introduced in \citet[Section 4.1.3]{kumar2012bootstrap}, the authors show that $C(X)$ converges in distribution to $\varepsilon(X)$ under mild assumptions on $f$ as the number of training samples and bootstraps tend to infinity. 

Thus, by adding the model predictions $f(X)$ to the values in $C(X)$, they get an estimation of the prediction distribution. 
Since convergence in distribution implies convergence of quantiles (see, e.g., Theorem 2A in \citet{parzen1980quantile}), this also implies that the resulting prediction intervals built using the quantiles of $C(X)$ are asymptotically correct. 

This theoretical justification leads us to consider a potential way to build a selective regressor, as sketched in Figure~\ref{fig:diagram-selective-regression}. Namely, we start by fitting the regression model over bootstrapped versions of the training set $(X^\text{train}, Y^\text{train})$ as in \citet{mougan2023monitoring}. Next, we consider the width of the estimated prediction intervals as our confidence function, and we calibrate a selection threshold $\tau$ on a separate calibration set during training. We refer to this method as \doubtint{}.
The intuition behind \doubtint{} is simple: the larger the interval, the more uncertain the prediction, hence abstention is preferable.

Since the optimal selective predictor (\ref{eq:optimal-selective-predictor}) thresholds the conditional variance function to build the selection function, we also consider a slight variation of \doubtint{} by directly using the variance of $C(X)$ rather than the width of the intervals.
We refer to this modification as \doubtvar{}.
Once again, the intuition is straightforward: the larger variance of the predictive distribution, the more uncertain the prediction is, making abstaining more likely.

We highlight that both \doubtint{} and \doubtvar{} are completely model-agnostic as they can be used with any off-the-shelf estimator.

\subsection{Explainable Selective Regression}

Using a confidence function to threshold a rejection of the model's predictions does not explain why either a given sample was rejected. We propose to solve this open issue by training a classifier to predict the accept/reject decisions of the original regressor model, after which we can perform a Shapley value analysis of the classifier, which can be used to explain why samples are being rejected or accepted. The process is pictured in Figure~\ref{fig:diagram-xai}. We start by splitting the data into four parts: training, calibration, validation and test. We use the training and calibration splits to train the selective regressor, as described in the previous subsection and shown in Figure~\ref{fig:diagram-selective-regression}. Next, we apply the associated selection function $s_f$ on the validation split $X^\text{validation}$ and fit an \textbf{audit model}
\begin{equation}
    \texttt{audit}\colon\mathcal{X}\to\{0,1\}
\end{equation}

on $(X^{\text{validation}}, s_f(X^{\text{validation}}))$ and compute the Shapley values \cite{shapTree} of $\texttt{audit}$ on the test split.

%% file: content/experiments.tex
\section{Experiments}

We present in this section the experiments to validate our proposed methodology, addressing the following questions:
\begin{itemize}

    \item [\textbf{Q1}] Does model-agnostic uncertainty estimation via non-parametric bootstrap achieve state-of-the-art performance for selective regression? 

    \item [\textbf{Q2}] Does the choice of different regression algorithms affect the results from \textbf{Q1}?

    
    \item [\textbf{Q3}] Can we characterize the features that led to the rejection decision of a single instance?
    
\end{itemize}

\subsection{Experimental Settings}

\paragraph{Data} For \textbf{Q1} and \textbf{Q2}, we consider 69 tabular regression datasets from existing regression benchmarks \cite{Olson2017PMLB,DBLP:conf/nips/GrinsztajnOV22}. Data include applications from different domains, such as finance, healthcare and natural sciences. 
To limit computational time, we exclude large datasets from the current study, i.e. $n \geq 100,000$, with $n$ representing the dataset size.
We additionally exclude datasets with a size $n \leq 100$, to have enough data to test different target coverages.
We further detail the datasets' characteristics in Table A1 of the appendix.
We apply one-hot encoding to categorical variables and employ feature and target variable Min-Max normalisation to account for the varying ranges in the feature and target distributions. 
We run experiments using five different seeds and consider the average results over these runs.

For \textbf{Q3}, we consider the popular House Prices regression dataset\footnote{\url{https://www.kaggle.com/c/house-prices-advanced-regression-techniques}} to illustrate how our explainable selective regression method works. The task is to predict the selling price of a given property, with a subset of the seven most predictive features.

\paragraph{Baselines}
As we employ tabular datasets, in which tree-based methods such as gradient boosting have been shown to achieve better performance \cite{DBLP:conf/nips/GorishniyRKB21, DBLP:conf/nips/GrinsztajnOV22}, we compare our approach to other non-deep learning model-agnostic methods from the literature. The baselines we consider are:

\textit{Plug-In}, a model-agnostic method introduced by \citet{DBLP:conf/nips/ZaouiDH20}. \plugin{} works as follows. Given a target coverage $\alpha$, training samples $(X^{\text{train}}, Y^{\text{train}})$ and \textit{unlabelled} calibration samples $X^{\text{cal}}\sim P_X$, it firstly estimates the selective regressor by fitting a regression function $\hat{f}$ on $(X^{\text{train}}, Y^{\text{train}})$.
Next, the training residuals $\varepsilon^\text{train} := (Y^\text{train}-\hat{f}(X^\text{train}))^2$ are computed. 
We then fit another regression function $\hat{g}$ on $(X^\text{train}, \varepsilon^\text{train})$ and compute the predicted residual values $\hat{g}(X^\text{train})$ of the calibration set $X^\text{cal}$.
Lastly, we use the predicted residuals to calibrate the selective regressor;

\textit{SCross}, an adaptation of the algorithm by ~\citet{PugnanaRuggieri2023a} to selective regression. \scross{} mitigates overfitting concerns of \plugin{} \cite{kennedy2020towards} by applying cross-validation to obtain validation residuals of $\hat{f}$ instead of training residuals. This is done by splitting the training set into $K$ folds, training the regressor on $K-1$ folds and computing residuals over the final $K$'th fold. We repeat this procedure $K$ times and use the residuals from all $K$ iterations to fit the $\hat{g}$ function. Finally, similarly to \plugin{}, \scross{} calibrates the selective regressor over the unlabeled set $X^\text{cal}$;
 
\textit{MAPIE}, a method that estimates the uncertainty around predictions using the conformal prediction technique from \citet{DBLP:conf/nips/KimXB20,DBLP:conf/nips/RomanoPC19,DBLP:conf/icml/Xu021}. We first fit the regressor over the training set to build the selective regressor. We then use the $\text{CV}^+$ technique \cite{barber2020predictive} with $K=5$ to produce prediction sets at 95\%, and we consider the width of the provided prediction interval as a proxy for the selection function $s$ (the larger the interval, the more likely the rejection). We finally calibrate the selective regressor over an unlabeled calibration dataset $X^\text{cal}$;

\textit{GoldCase}, an Oracle implementation with access to $Y$ labels. This method rejects instances whose residual value is above the $\alpha$-th percentile. \gold{} provides an upper bound to the performance of all the other baselines.

\paragraph{Hyperparameters} 
For \textbf{Q1}, we consider \texttt{XGBoost} \cite{DBLP:conf/kdd/ChenG16} as the base algorithm since it achieves state-of-the-art performance in many tasks \cite{DBLP:conf/nips/GrinsztajnOV22,DBLP:journals/corr/abs-2110-01889}. 

For \textbf{Q2}, we also consider \texttt{LightGBM} and \texttt{scikit-learn} implementations of \texttt{LinearRegression}, \texttt{DecisionTree} and \texttt{RandomForest}.
Hyperparameters are set to default API values.
For \doubtvar{} and \doubtint{}, we set the number of bootstraps to the default value $\sqrt{n_\text{train}}$, with $n_{\text{train}}$ representing the training set size.
For \scross{} and \mapie{}, we set the number of folds to their default value $K=5$.

\paragraph{Metrics}
For \textbf{Q1} and \textbf{Q2}, we compute actual coverage denoted as $\widehat{\texttt{cov}}(s)$, i.e. the sample counterpart of $\texttt{cov}(s)$. We use actual coverage to check whether existing methods satisfy the coverage constraint. Ideally, we want the difference $\texttt{cov}(s) - \widehat{\texttt{cov}}(s)$ to be negative; i.e., that the predicted coverage is no less than the target coverage. Since small violations could occur in practice, we consider as a coverage violation the following metric:

\begin{equation*}
    \texttt{CovSat}(s) := \indicator{}(\texttt{cov}(s) -\widehat{\texttt{cov}}(s)\geq \varepsilon)
\end{equation*}

where $\varepsilon$ is set to $0.05$ to account for reasonably small violations. 
We measure performance using the percentage decrease of Mean Squared Error, i.e.

\begin{equation*}
    \Delta \texttt{MSE}(h,g,c) := \frac{\texttt{MSE}(h,g,c)}{\texttt{MSE}(h,g,1)} -1,
\end{equation*}

where $\texttt{MSE}(h,g,c)$ is the Mean Squared Error computed over accepted instances at desired coverage $c$ and $ \texttt{MSE}(h,g,1)$ is the Mean Squared Error on the full sample. The lower $\Delta \texttt{MSE}$, the more the model can improve its performance once we allow abstention. This allows for a relative comparison of performances across multiple datasets and coverages. Moreover, since the lower the coverage, the lower the expected $\texttt{MSE}$, we need to account for coverage violations to guarantee a fair comparison across methods. Regarding this aspect, if $\texttt{CovSat}(s) = 0$, we set $\Delta \texttt{MSE} = 0$ to avoid rewarding those methods that improve performance by over-rejecting instances.

\paragraph{Hardware}
We use a 96 cores machine with Intel(R) Xeon(R) Gold 6342 CPU @ 2.80GHz and two NVIDIA RTX A6000, OS Ubuntu 20.04.4 for all the experiments.
The experiments took roughly two days on the server, with an estimated $CO_2$ consumption of $\sim 7.5kg$ according to \texttt{codecarbon} \cite{DBLP:journals/corr/abs-1910-09700,benoit_courty_2023_8181237}.

\subsection{Q1: Evaluating Bootstrap Uncertainty as a Confidence Function}
In this subsection, we provide results when evaluating \doubtint{} and \doubtvar{} as methods to perform selective regression.

\paragraph{Setup} 
We randomly split each dataset into a $60/20/20$ training/calibration/test split and train the selective regressor on the training set. We then choose different target coverages $c\in\{.99, .95, .90, .85, .80, .75, .70, .65, .60. .55, .50\}$, and calibrate a selection function over the calibration set for each of these. Lastly, we compute $\texttt{CovSat}$ and $\Delta\texttt{MSE}$ of the calibrated selection functions over accepted instances on the test set.

\paragraph{Results}

\begin{figure*}
    \centering
    
    \begin{subfigure}[b]{.31\textwidth} 
        \includegraphics[scale=.18]{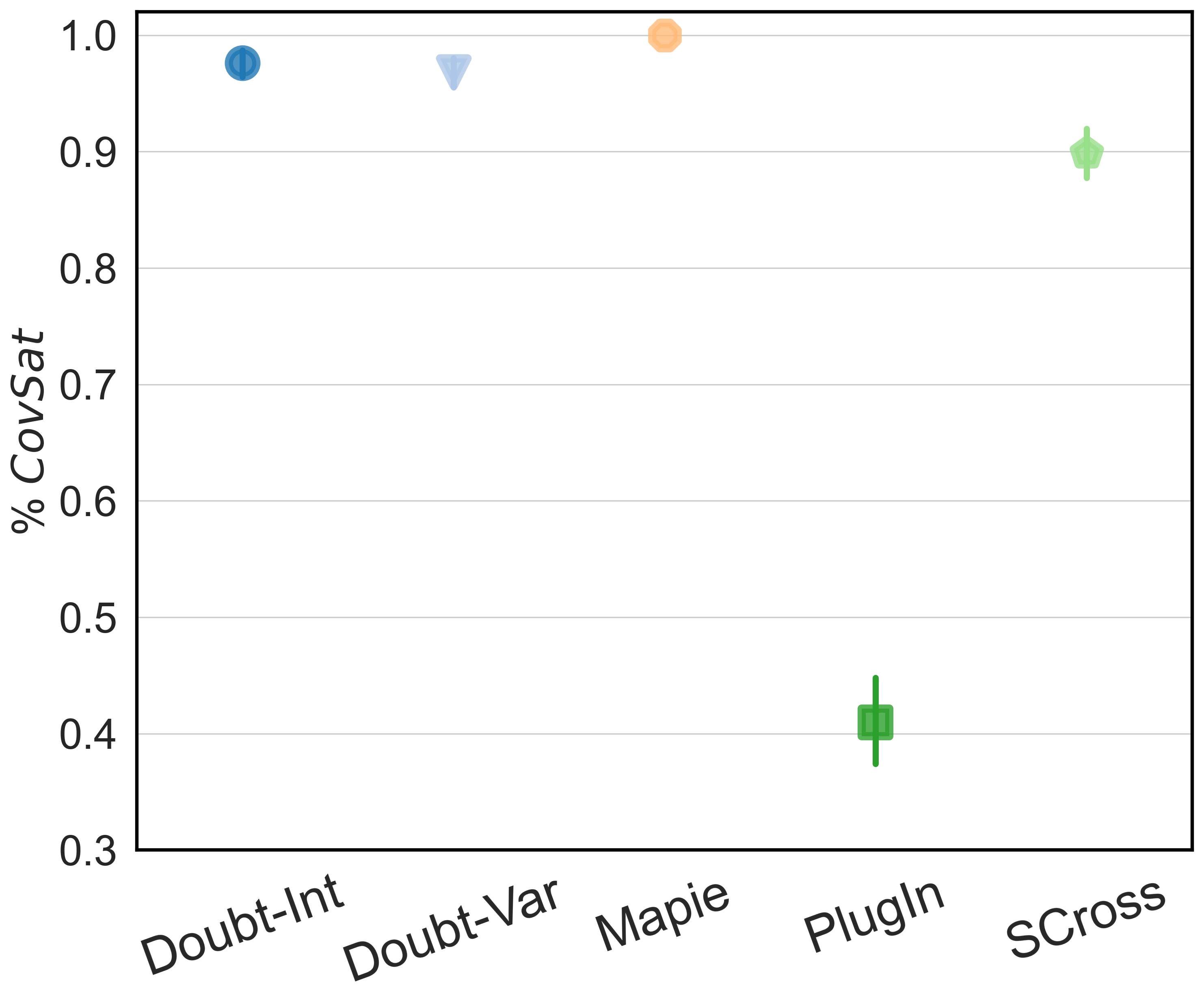}
        \caption{\% $\texttt{CovSat}=1$ for different baselines (higher is better)}
        \label{fig:pmlbMSE}
    \end{subfigure}
    \hfill
    \begin{subfigure}[b]{.31\textwidth} 
        \includegraphics[scale=.18]{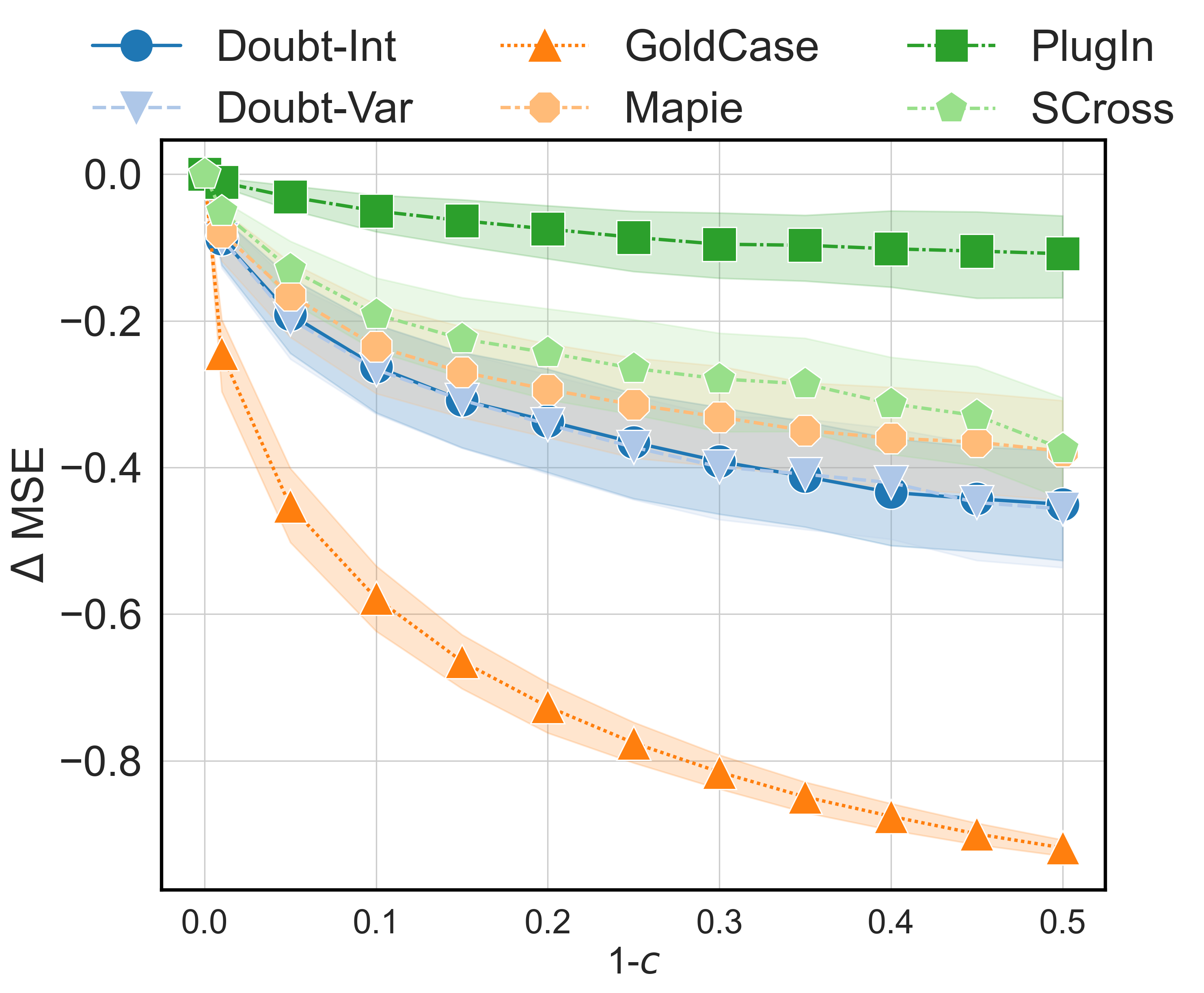}
        \caption{$\Delta \texttt{MSE}$ for reject rate ($1-c$) levels (lower is better).}
        \label{fig:CDplot.75}
    \end{subfigure}
    \hfill
    \begin{subfigure}[b]{.31\textwidth} 
        \includegraphics[scale=.2]{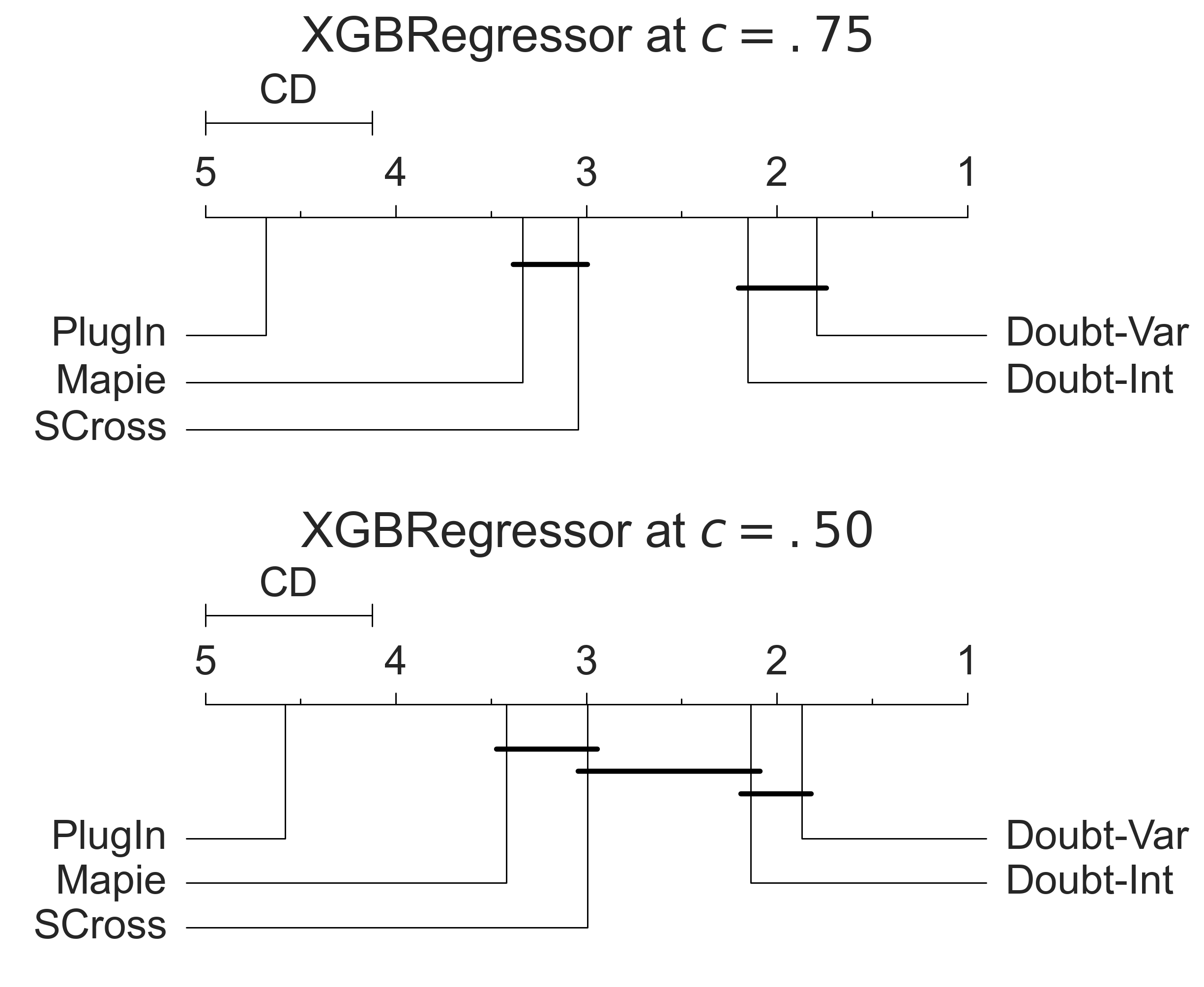}
        \caption{CD plots at $c=.75$ and $c=.50$.}
        \label{fig:CDplot.50}
    \end{subfigure}
    
    
    \caption{ The \textbf{Q1} experiments, showing the results on selected datasets using \texttt{XGBoost}. In the Critical Difference plot in \textbf{(c)}, the differences are statistically significant whenever the bold lines do not intersect multiple baselines.}
\end{figure*}


Figure~\ref{fig:pmlbMSE} reports the percentage of times the baselines achieve $\texttt{CovSat} = 1$ over our experiments when using \texttt{XGBoost} as the base regressor. We can see that the only method with large coverage violations is \plugin{}, with a percentage of $\sim41\%$ of the experiments. This happens as the selective regressor overfit data, failing to properly estimate the conditional variance function. 
A large part of coverage violations occurs on small datasets ($n<1,000$), where \plugin{} achieves desired coverage only $8.8\%$ of the time. On the other hand, all the other methods show a percentage of roughly $\sim90\%$, with \mapie{} achieving $100\%$ of satisfied constraints.

Figure~\ref{fig:CDplot.75} shows the average $\Delta \texttt{MSE}$ values for the different baseline methods and coverages over the analyzed datasets. Due to space limits, we report results at dataset level in the appendix.
For lower coverages, \doubtint{} and \doubtvar{} reach the best performance, with an average drop in MSE of $\sim.45\%$ and $\sim45.6\%$, respectively, at $c=.5$.
\plugin{} achieves the worst performance, with an average drop at $c=.5$ of $\sim10.8\%$, which is more than three times less the drop achieved by \doubtvar{} and \doubtint{}.

To evaluate the statistical significance of our results, we use the Nemenyi post-hoc test at a 95\% significance level ~\cite{DBLP:journals/jmlr/Demsar06}. Figure~\ref{fig:CDplot.50} provides the Critical Difference (CD) plots resulting from the tests at $c=.75$ and $c=.5$ (the median and minimum values of target coverages). 
At $c=.75$, the difference between the bootstrap uncertainty estimation strategies and all the other baselines is statistically significant. On the other hand, the performance of \doubtvar{} and \doubtint{} is not statistically distinguishable. 
This result also holds when decreasing the coverage further, with \doubtvar{} and \doubtint{} outperforming other methods and still not different from each other in a statistically significant sense.
Hence, experimental evidence seems to support the usage of bootstrap uncertainty estimation as a reliable way to perform selective regression, answering \textbf{Q1} in the positive.

\subsection{Q2: Evaluating the Regressor Choice}
In this subsection, we investigate how the choice of the base regressor affects the results of bootstrap-based methods and other model-agnostic baselines.

\begin{figure*}[t]
    \centering

    \begin{subfigure}[b]
        {.33\textwidth} 
        \centering
        \includegraphics[scale=.185]{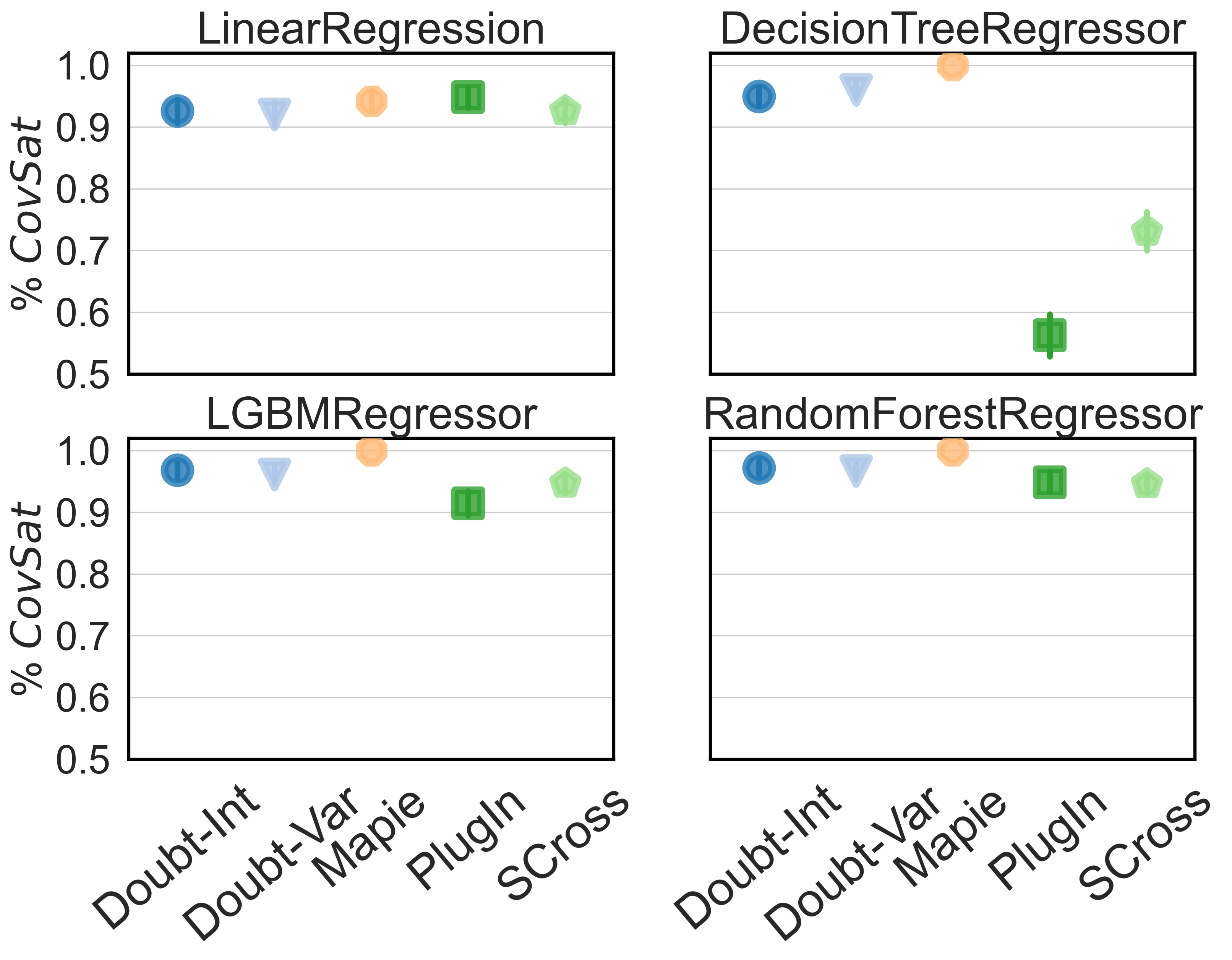}
        \caption{\% $\texttt{CovSat}=1$ for different regressors.}
        \label{fig:CDplotLGBM}
    \end{subfigure}
    \hfill
    \begin{subfigure}[b]{.62\textwidth} 
        \includegraphics[scale=.205]{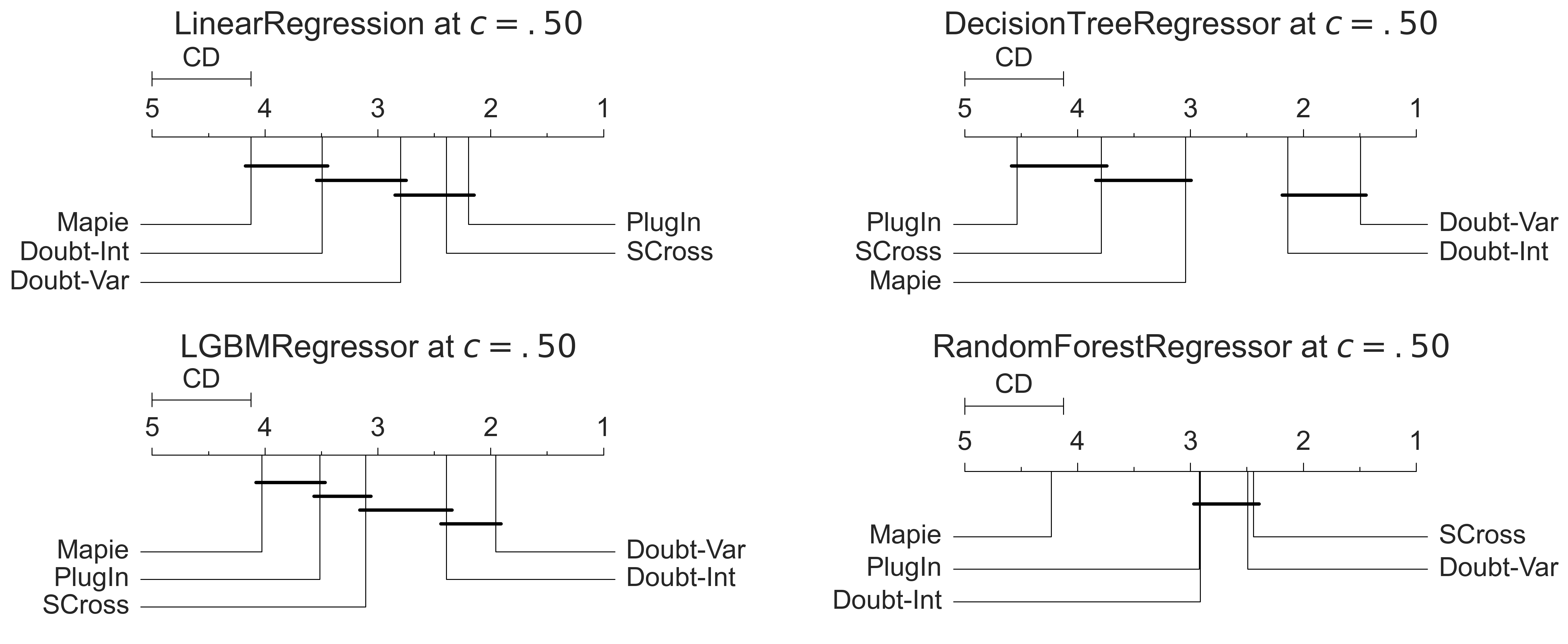}
        \caption{CD plot for different regressors.}
        \label{fig:CDplotDT}
    \end{subfigure}

\caption{The \textbf{Q2} experiments, comparing the performance when varying the underlying regression algorithm.}
\end{figure*}

\paragraph{Setup} We follow the same setup of \textbf{Q1}, and consider the base regressors presented in the hyperparameters paragraph.

\paragraph{Results} Figure~\ref{fig:CDplotLGBM},  displays the percentage of $\texttt{CovSat}=1$ for the different algorithms analyzed. We see that the only algorithm where large coverage violations occur is \texttt{DecisionTree}, where \plugin{} and \scross{} satisfy coverage constraints on $\sim 56.3$\% and $\sim 73.1$\% of the experiments, respectively.
The coverage is satisfied for all the other regression algorithms in roughly $90$\% of the cases.

To evaluate the robustness of bootstrap-based methods, we run the Nemenyi post-hoc test using $\Delta \texttt{MSE}$ for the different regression algorithms, and we report the CD plots at the minimal coverage $c=.5$ in Figure~\ref{fig:CDplotDT}.
When considering \texttt{LinearRegression}, \plugin{} is ranked first, even though with no statistically significant difference with respect to \scross{} and \doubtvar{}. 
When focusing on the \texttt{DecisionTree} regressor, \doubtvar{} and \doubtint{} are ranked first and second, respectively,
with statistically significant differences from the other baselines not based on bootstrap uncertainty estimation. 
 
When evaluating \texttt{LightGBM}, \doubtvar{} and \doubtint{} achieve the top-2 positions. However, \doubtint{} is not statistically different from \scross{}.  
On the other hand, when considering \texttt{RandomForest}, all the methods, aside from MAPIE, are indistinguishable.

These results suggest that our bootstrap-based approaches help improve performance.
In particular, our method is preferable when we consider regression algorithms prone to overfitting, such as single decision trees. 
This is because \plugin{} learns the selection function on the same data used to build the regression function, leading to potential over-fitting concerns \cite{kennedy2020towards}.
Thus, the findings of our analysis constitute a negative answer to \textbf{Q2}, since \doubtvar{} is always as good or better than previous SOTA methods.

\subsection{Q3: Explainable Model Agnostic Selective Regression}

In this section, we show how our methodology of explainable selective regression, as illustrated in Figure 2, can assist users in auditing rejection decisions.

\paragraph{Setup}

Since this section aims to show how we can characterize the selection strategy, we consider an illustrative setup inspired by previous work on model degradation by ~\cite{mougan2023monitoring}. 
Given a dataset, we add a randomly generated feature $X_{Random}$ to generate a feature independent of target variable $Y$.
We then randomly split data according to a  25/25/25/25 proportion between the training, calibration, validation and test set.
We use training and calibration sets to learn a \doubtvar{} selective regressor on the extended feature space, using \texttt{XGBoost} algorithm as the base model. 
We train an \texttt{audit} model on the validation set, following the steps described in the methodology section. To build the \texttt{audit} model, we employ the default \texttt{sklearn} implementation of \texttt{LogisticRegression} and set a target coverage $c=.80$. Due to space limits, we provide results for different algorithm choices in Appendix Table A2.

Subsequently, we induce a distribution shift to force the model into rejecting samples that would have otherwise been accepted. To achieve this, we isolate the accepted instances from the test set. Then, iteratively, for every feature within the expanded feature space, we perturb the variable distribution through Gaussian noise $\varepsilon\sim\mathcal{N}(5, 1)$.

We use the full conditional SHAP technique to estimate the Shapley values distribution of the \texttt{audit} model for both the perturbed and the non-perturbed case. This is done because such a technique respects the correlations among the input features, so if the model depends on one input and that input is correlated with another, both get some credit for the model’s behavior~\cite{DBLP:conf/nips/LundbergL17,lundberg2018explainable}. 
We then compute the Wasserstein distance~\cite{kantorovich1960mathematical,vaserstein1969markov} between the univariate Shapley value distribution of the \texttt{audit} model in both non-perturbated and perturbated cases to measure how much the perturbations affect explanations distributions.
We finally consider perturbations occurring at the same time on $X_{Random}$ and $X_{GrLivArea}$ (one of the most predictive features) to show how one can characterize the selection function through a visualization of the explanations distributions.

\paragraph{Results}

\begin{figure}[h!] 
  \centering
  \includegraphics[width=1\linewidth]{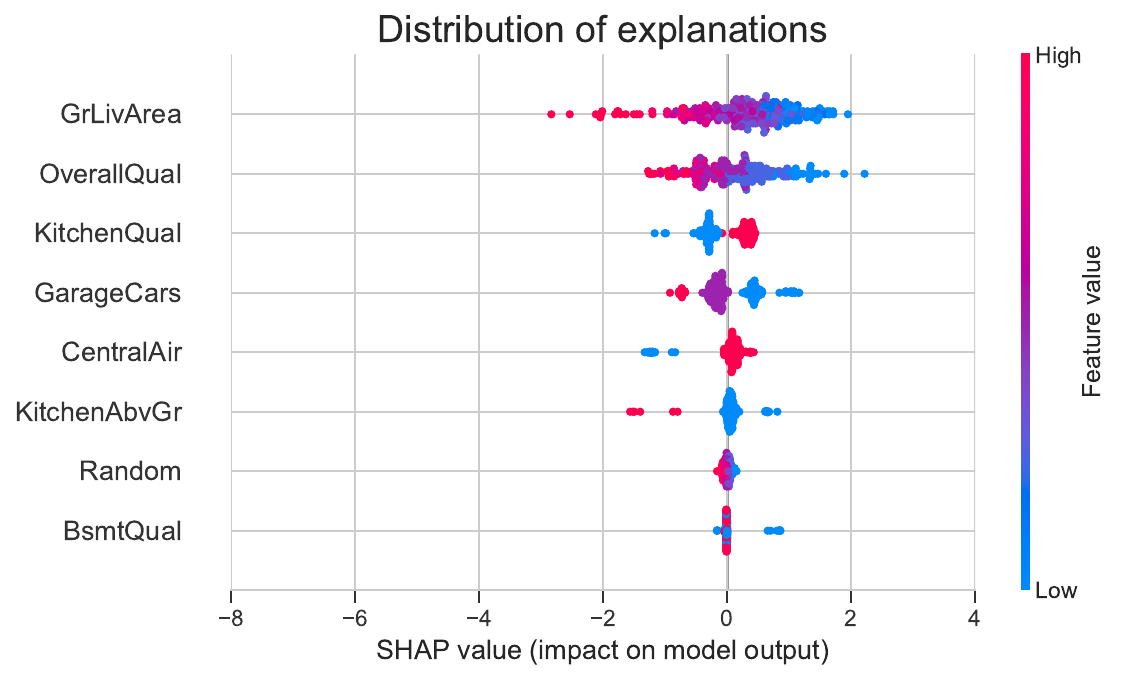}
  \includegraphics[width=1\linewidth]{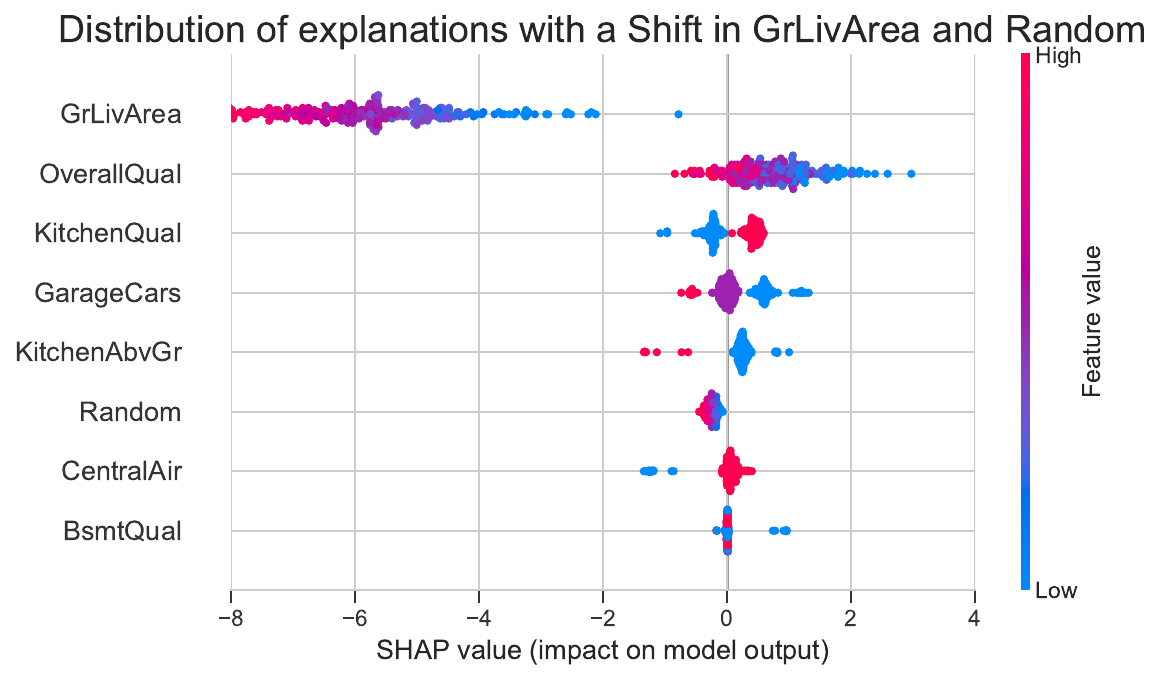}
  \caption{The plot illustrates two features, $X_{\text{Random}}$ and $X_{\text{GrLiveArea}}$, of the House Regression dataset, being simultaneously shifted. 
  The top image displays local explanations for the original distributions, while the bottom image shows the shifted features.}
  \label{fig:loca_xai}
\end{figure}

\begin{table}[h!]
\centering
\begin{tabular}{c|c}
\textbf{Feature}        & \textbf{Wasserstein Distance} \\ \hline
GrLivArea      & $5.38\pm0.06$                 \\
OverallQual    & $3.83\pm0.03$                 \\
CentralAir     & $2.32\pm0.02$                 \\
KitchenAbvGr   & $1.75\pm0.02$                 \\
BsmtQual       & $0.76\pm0.01$                 \\
KitchenQual    & $0.66\pm0.01$                 \\
GarageCars     & $0.463\pm0.003$                 \\
Random         & $0.175\pm0.001$                
\end{tabular}
\caption{Wasserstein distance over the Shapley distributions of the predicted samples when univariate shifting the features for the House Regression dataset (higher is a more impactful shift). Results are reported as the average $\pm$ standard deviation over different runs.}
\end{table}

Table 1 provides the Wasserstein distance obtained when shifting the features individually. We can see that the distribution of the explanations for the random variable $X_{random}$ is less affected by the perturbation, as low Wasserstein implies more similar distributions.
On the contrary, the Shapley distribution of the selection function explanations shifts more substantially for features with more predictive power to determine the acceptance/rejection.

We further illustrate the mechanism behind the selective function in Figure 5, where we plot the Shapley distributions when considering a shift occurring simultaneously in $X_{Random}$ and $X_{GrLiveArea}$. The top image represents the Shapley values distribution when no shift has occurred over the accepted instances, while the bottom image depicts the distribution in the shifted scenario. We notice that the perturbation in $X_{GrLiveArea}$ shifts Shapley distribution towards negative values, decreasing the chances for instances to be accepted by the final selective regressor. This aligns with our desiderata, as we would like to abstain more when evaluating instances belonging to the feature space's unexplored (hence more uncertain) areas. On the other hand, the shift in $X_{\text{Random}}$ has a less impact on the distribution of the Shapley values, showing that non-relevant features shift do not affect the acceptance/rejection decision. Thus, we can positively answer to \textbf{Q3}, as our explainable selective regression methodology allows us to describe the key drivers of the selection function. 

%% file: content/conclusions.tex
\section{Conclusions}
This paper addresses the understudied problem of selective regression, introducing a novel method based on the state-of-the-art uncertainty estimation technique by \citet{mougan2023monitoring}. This allows any regressor to abstain from making predictions when facing high uncertainty. Our method is theoretically grounded and completely model-agnostic. We provide an extensive empirical evaluation over 69 datasets and different regression algorithms, showing how our approach is as good or better than competing methods and is especially useful when over-fitting concerns might occur. Finally, we use
Shapley values, in conjunction with the proposed uncertainty-based selective regressor, to identify which features are driving rejection, providing a simple-yet-effective method to validate the selection function.


\paragraph{Limitations:} 
We confine our analysis to datasets with sizes ranging from $100 < n < 100,000$. Within this range, our techniques exhibit enhancements over prevailing methods across a spectrum of base algorithms. Moreover, our benchmark exclusively encompasses non-deep learning methods, which facilitates expedited algorithm training through bootstrap samples. The extension of our approach to larger deep-learning models could entail heightened complexity due to potential computational intensiveness.
Furthermore, our explainable AI approach rests on the estimation of Shapley values, and varying choices in models or Shapley value approximations may yield disparate outcomes.

\paragraph{Reproducibility Statement}\label{sec:reproducibility}
To ensure reproducibility of our results, we make the data, data preparation routines, code repositories, and methods publicly available\footnote{\url{https://anonymous.4open.science/r/SelectiveRegression-282E/README.md}}. 
Our methods will be included in the open-source Python package \texttt{doubt}.

%% file: appendixV2.tex
\title{Supplementary Material for Model Agnostic Explainable Selective Regression via Uncertainty Estimation}
\maketitle
\section{Supplementary Material for Model Agnostic Explainable Selective Regression via Uncertainty Estimation}
\subsection{Data Details}
We provide in Table A1 the set of data employed in our analysis. We report the name of the dataset, the link to the original repository, the number of training instances and the number of features employed in the regression task.

\begin{table*}[h!]
\centering
\caption{Dataset Details}
\label{tab:data2}
\resizebox{\textwidth}{!}{
\begin{tabular}{cccc}
\toprule
                 \textbf{Dataset} &                                                                            \textbf{Link} & \textbf{Training Size} &  \textbf{Feature Space Dim.} \\
\midrule
                        1027\_ESL &                         \url{https://epistasislab.github.io/pmlb/profile/1027\_ESL.html} &                    292 &                            4 \\
                        1028\_SWD &                         \url{https://epistasislab.github.io/pmlb/profile/1028\_SWD.html} &                    600 &                           10 \\
                        1029\_LEV &                         \url{https://epistasislab.github.io/pmlb/profile/1029\_LEV.html} &                    600 &                            4 \\
                        1030\_ERA &                         \url{https://epistasislab.github.io/pmlb/profile/1030\_ERA.html} &                    600 &                            4 \\
                1193\_BNG\_lowbwt &                 \url{https://epistasislab.github.io/pmlb/profile/1193\_BNG\_lowbwt.html} &                 18,662 &                            9 \\
            1199\_BNG\_echoMonths &             \url{https://epistasislab.github.io/pmlb/profile/1199\_BNG\_echoMonths.html} &                 10,497 &                            9 \\
                    197\_cpu\_act &                     \url{https://epistasislab.github.io/pmlb/profile/197\_cpu\_act.html} &                  4,915 &                           21 \\
                         201\_pol &                          \url{https://epistasislab.github.io/pmlb/profile/201\_pol.html} &                  9,000 &                           48 \\
                   207\_autoPrice &                    \url{https://epistasislab.github.io/pmlb/profile/207\_autoPrice.html} &                     95 &                           15 \\
                       210\_cloud &                        \url{https://epistasislab.github.io/pmlb/profile/210\_cloud.html} &                     64 &                            5 \\
                    215\_2dplanes &                     \url{https://epistasislab.github.io/pmlb/profile/215\_2dplanes.html} &                 24,460 &                           10 \\
                   218\_house\_8L &                    \url{https://epistasislab.github.io/pmlb/profile/218\_house\_8L.html} &                 13,670 &                            8 \\
                     225\_puma8NH &                      \url{https://epistasislab.github.io/pmlb/profile/225\_puma8NH.html} &                  4,915 &                            8 \\
                  227\_cpu\_small &                   \url{https://epistasislab.github.io/pmlb/profile/227\_cpu\_small.html} &                  4,915 &                           12 \\
                    229\_pwLinear &                     \url{https://epistasislab.github.io/pmlb/profile/229\_pwLinear.html} &                    120 &                           10 \\
                230\_machine\_cpu &                 \url{https://epistasislab.github.io/pmlb/profile/230\_machine\_cpu.html} &                    125 &                            6 \\
            294\_satellite\_image &             \url{https://epistasislab.github.io/pmlb/profile/294\_satellite\_image.html} &                  3,861 &                           36 \\
                          344\_mv &                           \url{https://epistasislab.github.io/pmlb/profile/344\_mv.html} &                 24,460 &                           10 \\
4544\_GeographicalOriginalofMusic & \url{https://epistasislab.github.io/pmlb/profile/4544\_GeographicalOriginalofMusic.html} &                    635 &                          117 \\
                        503\_wind &                         \url{https://epistasislab.github.io/pmlb/profile/503\_wind.html} &                  3,944 &                           14 \\
                     505\_tecator &                      \url{https://epistasislab.github.io/pmlb/profile/505\_tecator.html} &                    144 &                          124 \\
                      519\_vinnie &                       \url{https://epistasislab.github.io/pmlb/profile/519\_vinnie.html} &                    228 &                            2 \\
                        522\_pm10 &                         \url{https://epistasislab.github.io/pmlb/profile/522\_pm10.html} &                    300 &                            7 \\
                      529\_pollen &                       \url{https://epistasislab.github.io/pmlb/profile/529\_pollen.html} &                  2,308 &                            4 \\
                      537\_houses &                       \url{https://epistasislab.github.io/pmlb/profile/537\_houses.html} &                 12,384 &                            8 \\
                         547\_no2 &                          \url{https://epistasislab.github.io/pmlb/profile/547\_no2.html} &                    300 &                            7 \\
         556\_analcatdata\_apnea2 &          \url{https://epistasislab.github.io/pmlb/profile/556\_analcatdata\_apnea2.html} &                    285 &                            3 \\
         557\_analcatdata\_apnea1 &          \url{https://epistasislab.github.io/pmlb/profile/557\_analcatdata\_apnea1.html} &                    285 &                            3 \\
                     560\_bodyfat &                      \url{https://epistasislab.github.io/pmlb/profile/560\_bodyfat.html} &                    151 &                           14 \\
                       564\_fried &                        \url{https://epistasislab.github.io/pmlb/profile/564\_fried.html} &                 24,460 &                           10 \\
                  574\_house\_16H &                   \url{https://epistasislab.github.io/pmlb/profile/574\_house\_16H.html} &                 13,670 &                           16 \\
            581\_fri\_c3\_500\_25 &             \url{https://epistasislab.github.io/pmlb/profile/581\_fri\_c3\_500\_25.html} &                    300 &                           25 \\
            582\_fri\_c1\_500\_25 &             \url{https://epistasislab.github.io/pmlb/profile/582\_fri\_c1\_500\_25.html} &                    300 &                           25 \\
            584\_fri\_c4\_500\_25 &             \url{https://epistasislab.github.io/pmlb/profile/584\_fri\_c4\_500\_25.html} &                    300 &                           25 \\
           586\_fri\_c3\_1000\_25 &            \url{https://epistasislab.github.io/pmlb/profile/586\_fri\_c3\_1000\_25.html} &                    600 &                           25 \\
           589\_fri\_c2\_1000\_25 &            \url{https://epistasislab.github.io/pmlb/profile/589\_fri\_c2\_1000\_25.html} &                    600 &                           25 \\
           592\_fri\_c4\_1000\_25 &            \url{https://epistasislab.github.io/pmlb/profile/592\_fri\_c4\_1000\_25.html} &                    600 &                           25 \\
           598\_fri\_c0\_1000\_25 &            \url{https://epistasislab.github.io/pmlb/profile/598\_fri\_c0\_1000\_25.html} &                    600 &                           25 \\
            605\_fri\_c2\_250\_25 &             \url{https://epistasislab.github.io/pmlb/profile/605\_fri\_c2\_250\_25.html} &                    150 &                           25 \\
           620\_fri\_c1\_1000\_25 &            \url{https://epistasislab.github.io/pmlb/profile/620\_fri\_c1\_1000\_25.html} &                    600 &                           25 \\
            633\_fri\_c0\_500\_25 &             \url{https://epistasislab.github.io/pmlb/profile/633\_fri\_c0\_500\_25.html} &                    300 &                           25 \\
            643\_fri\_c2\_500\_25 &             \url{https://epistasislab.github.io/pmlb/profile/643\_fri\_c2\_500\_25.html} &                    300 &                           25 \\
            644\_fri\_c4\_250\_25 &             \url{https://epistasislab.github.io/pmlb/profile/644\_fri\_c4\_250\_25.html} &                    150 &                           25 \\
            653\_fri\_c0\_250\_25 &             \url{https://epistasislab.github.io/pmlb/profile/653\_fri\_c0\_250\_25.html} &                    150 &                           25 \\
            658\_fri\_c3\_250\_25 &             \url{https://epistasislab.github.io/pmlb/profile/658\_fri\_c3\_250\_25.html} &                    150 &                           25 \\
                   663\_rabe\_266 &                    \url{https://epistasislab.github.io/pmlb/profile/663\_rabe\_266.html} &                     72 &                            2 \\
            665\_sleuth\_case2002 &             \url{https://epistasislab.github.io/pmlb/profile/665\_sleuth\_case2002.html} &                     88 &                            6 \\
              666\_rmftsa\_ladata &               \url{https://epistasislab.github.io/pmlb/profile/666\_rmftsa\_ladata.html} &                    304 &                           10 \\
  678\_visualizing\_environmental &   \url{https://epistasislab.github.io/pmlb/profile/678\_visualizing\_environmental.html} &                     66 &                            3 \\
         690\_visualizing\_galaxy &          \url{https://epistasislab.github.io/pmlb/profile/690\_visualizing\_galaxy.html} &                    193 &                            4 \\
                695\_chatfield\_4 &                 \url{https://epistasislab.github.io/pmlb/profile/695\_chatfield\_4.html} &                    141 &                           12 \\
            712\_chscase\_geyser1 &             \url{https://epistasislab.github.io/pmlb/profile/712\_chscase\_geyser1.html} &                    133 &                            2 \\
                          abalone &                     \url{https://www.openml.org/search?type=data&status=active&id=42726} &                  2,506 &                            9 \\
                            bikes &                     \url{https://www.openml.org/search?type=data&status=active&id=42712} &                 10,427 &                           16 \\
                brazilian\_houses &                     \url{https://www.openml.org/search?type=data&status=active&id=42688} &                  6,415 &                           48 \\
                         diamonds &                     \url{https://www.openml.org/search?type=data&status=active&id=42225} &                 32,364 &                           23 \\
                        elevators &                       \url{https://www.openml.org/search?type=data&status=active&id=216} &                  9,959 &                           18 \\
                     house\_sales &                     \url{https://www.openml.org/search?type=data&status=active&id=42731} &                 12,967 &                           21 \\
                         mercedes &                     \url{https://www.openml.org/search?type=data&status=active&id=42570} &                  2,525 &                          555 \\
                            miami &                     \url{https://www.openml.org/search?type=data&status=active&id=43093} &                  8,359 &                           16 \\
                     nikuradse\_1 &                      \url{https://epistasislab.github.io/pmlb/profile/nikuradse\_1.html} &                    217 &                            2 \\
                     nikuradse\_2 &                      \url{https://epistasislab.github.io/pmlb/profile/nikuradse\_2.html} &                    217 &                            1 \\
                          seattle &                     \url{https://www.openml.org/search?type=data&status=active&id=42496} &                 31,414 &                          293 \\
                             soil &                       \url{https://www.openml.org/search?type=data&status=active&id=688} &                  5,184 &                            4 \\
                           sulfur &                     \url{https://www.openml.org/search?type=data&status=active&id=23515} &                  6,048 &                            5 \\
                     superconduct &                     \url{https://www.openml.org/search?type=data&status=active&id=43174} &                 12,757 &                           81 \\
                          supreme &                       \url{https://www.openml.org/search?type=data&status=active&id=504} &                  2,431 &                            7 \\
                           topo21 &                       \url{https://www.openml.org/search?type=data&status=active&id=422} &                  5,331 &                          266 \\
                          y\_prop &                       \url{https://www.openml.org/search?type=data&status=active&id=416} &                  5,331 &                          251 \\
\bottomrule
\end{tabular}
}
\end{table*}

\subsection{Additional Results for Q3}
To show how the performance of the \texttt{audit} model depends on the base regressor and the classifier algorithm employed, Table A2 reports the Area Under the ROC Curve (AUC) score on the test set achieved by different \texttt{audit} models. The results show that the best fit is achieved when pairing $\texttt{XGBoost}$ with a \texttt{LogisticRegrssion}. At the same time, the worst performance is obtained when using the \texttt{LinearRegression} algorithm and the \texttt{LogisticRegression}.


\begin{table*}[h!]
\centering
\caption{AUC obtained by the \texttt{audit} model on predicting which instances will be rejected by \doubtvar{} on the House Regression dataset.}
\resizebox{\textwidth}{!}{
\begin{tabular}{l|cccc}
\multicolumn{1}{c|}{}           & \multicolumn{4}{c}{\textbf{Estimator}}                                                                    \\ \hline
\textbf{Auditor}                & \textbf{\texttt{DecisionTree}} & \textbf{\texttt{XGBoost}} & \textbf{\texttt{LinearRegression}} & \textbf{\texttt{RandomForest}} \\ \hline
\texttt{DecisionTreeClassifier} & 0.872                 & 0.892                         & 0.654                     & 0.872                 \\
\texttt{XGBoostClassifier}       & 0.888                 & 0.902                         & 0.671                     & 0.896                 \\
\texttt{KNeighborsClassifier}   & 0.868                 & 0.891                         & 0.694                     & 0.917                 \\
\texttt{LogisticRegression}     & 0.897                 & 0.935                         & 0.609                     & 0.884                 \\
\texttt{MLPClassifier}          & 0.903                 & 0.876                         & 0.673                     & 0.889                 \\
\texttt{RandomForestClassifier}           & 0.86                  & 0.91                          & 0.681                     & 0.884                
\end{tabular}
}

\end{table*}